\def\year{2022}\relax
\documentclass[letterpaper]{article} 
\usepackage{aaai22}  
\usepackage{times}  
\usepackage{helvet}  
\usepackage{courier}  
\usepackage[hyphens]{url}  
\usepackage{graphicx} 
\usepackage{placeins} 
\usepackage{natbib}  
\usepackage{caption} 
\DeclareCaptionStyle{ruled}{labelfont=normalfont,labelsep=colon,strut=off} 
\usepackage{algorithm}
\usepackage{algorithmic}
\usepackage{color,soul}

\usepackage{newfloat}
\usepackage{listings}
\floatstyle{ruled}
\newfloat{listing}{tb}{lst}{}
\floatname{listing}{Listing}
\usepackage{xcolor}
\usepackage{amsmath,amssymb}
\usepackage{latexsym}
\usepackage{paralist}
\usepackage{booktabs}
\usepackage{multirow}
\usepackage{tabularx}

\usepackage{cancel}

\def\argmax{\arg\,\max}

\title{DetIE: Multilingual Open Information Extraction \\ Inspired by Object Detection}
\author {
    Michael Vasilkovsky\textsuperscript{\rm 1,10},
    Anton Alekseev \textsuperscript{\rm 3,4},
    Valentin Malykh \textsuperscript{\rm 2,3,6,9},
    Ilya Shenbin \textsuperscript{\rm 3}, \\
    Elena Tutubalina \textsuperscript{\rm 5,6,7},
    Dmitriy Salikhov \textsuperscript{\rm 7},
    Mikhail Stepnov \textsuperscript{\rm 7},
    Andrey Chertok \textsuperscript{\rm 7,8}, \\
    Sergey Nikolenko \textsuperscript{\rm 3,9,10}
}
\affiliations {
    \textsuperscript{\rm 1} Skolkovo Institute of Science and Technology, Moscow, Russia\\
    \textsuperscript{\rm 2} Huawei Noah's Ark lab, Moscow, Russia \\
    \textsuperscript{\rm 3} St. Petersburg Department
of Steklov Mathematical Institute
of Russian Academy of Sciences, St. Petersburg, Russia\\
    \textsuperscript{\rm 4} St. Petersburg State University, St. Petersburg, Russia \\
    \textsuperscript{\rm 5} HSE University, Moscow, Russia\\
    \textsuperscript{\rm 6} Kazan Federal University, Kazan, Russia\\
    \textsuperscript{\rm 7} Sber AI, Moscow, Russia\\
    \textsuperscript{\rm 8} Artificial Intelligence Research Institute, Moscow, Russia\\
    \textsuperscript{\rm 9} ISP RAS Research Center for Trusted Artificial Intelligence, Moscow, Russia\\
    \textsuperscript{\rm 10} Neuromation OU, Tallinn, Estonia \\
}

\newcommand{\added}[1]{\textcolor{black}{#1}}
\newcommand{\camera}[1]{\textcolor{black}{#1}}

\begin{document}
\maketitle

\begin{abstract}
State of the art neural methods for open information extraction (OpenIE) usually extract triplets (or tuples) iteratively in an autoregressive or predicate-based manner in order not to produce duplicates. In this work, we propose a different approach to the problem that can be equally or more successful. Namely, we present a novel single-pass method for OpenIE inspired by object detection algorithms from computer vision. We use an order-agnostic loss based on bipartite matching that forces unique predictions and a Transformer-based encoder-only architecture for sequence labeling. The proposed approach is faster and shows superior or similar performance in comparison with state of the art models on standard benchmarks in terms of both quality metrics and inference time. Our model sets the new state of the art performance of 67.7\% F1 on CaRB evaluated as OIE2016 while being 3.35x faster at inference \added{than previous state of the art}. We also evaluate the multilingual version of~our model in~the zero-shot setting for two languages and introduce a~strategy for generating synthetic multilingual data to fine-tune the model for each specific language. In this setting, we show performance improvement \added{of 15\%} on multilingual Re-OIE2016, reaching 75\%~F1 for both Portuguese and Spanish languages. Code and models are available at \url{https://github.com/sberbank-ai/DetIE}.
\end{abstract}

\section{Introduction}\label{sec:introduction}
Extracting structured information from raw texts is a key area of research in natural language processing (NLP). It has a core set of well-defined basic problems: relation extraction, named entity recognition (NER), slot filling, and so on, each defining a specific view on the perception and analysis of textual data. 
In this work, we follow the paradigm of open information extraction (OpenIE)
that represents texts from an arbitrary domain as a set of (subject, relation, object) triplets~\cite{yates-etal-2007-textrunner}.
OpenIE methods do not rely on a pre-defined ontology schema and are trained to be domain-agnostic, so they can be used in many downstream NLP tasks: multi-document question answering and summarization \cite{DBLP:journals/corr/abs-1910-08435}, event schema induction \cite{balasubramanian-etal-2013-generating}, fact salience \cite{ponza-etal-2018-facts}, word embedding generation \cite{stanovsky-etal-2015-open}, and more. Using triplets as graph edges, OpenIE systems serve as a core component for unsupervised knowledge graph construction~\cite{Mausam2016OpenIE}; high-quality OpenIE systems can output an open knowledge graph even without further post-processing.

Historically, OpenIE systems were purely statistical or rule-based, often consisting of several components such as PoS (part-of-speech) tagging or syntax parsing, so errors tended to accumulate. Recently, end-to-end neural network systems for OpenIE have begun to outperform their non-neural counterparts.  There exist two paradigms in neural OpenIE: \emph{sequence labeling}~\cite{stanovsky-etal-2018-supervised,roy-etal-2019-supervising,kolluru2020openie6} and \emph{sequence generation} \cite{cui-etal-2018-neural,kolluru2020imojie}, each with their own merits and drawbacks.
To avoid duplicates, in both paradigms triplets are usually extracted iteratively either in the autoregressive \cite{kolluru2020openie6,kolluru2020imojie} or predicate-based manner \cite{ro2020multi}.
During training, autoregressive methods predict triplets in a prefedined order that usually has no meaning and excessively penalizes the model.
Predicate-based methods first extract all predicates and then iteratively find arguments for each, assuming that a predicate occurs in only one chain of arguments, which may not hold both in common benchmarks and in the real world.

In this work, we view OpenIE from a different perspective, as a direct set prediction problem. Our approach is inspired by one-stage anchor-based object detection models from computer vision~\cite{SSD,tan2020efficientdet}
that predict all bounding boxes in one forward pass and apply intersection-based matching to match predictions with the ground truth.
We bridge the inter-discipline gap and bring this idea to OpenIE, demonstrating increased or equivalent performance compared to state of the art methods.

We train our model on two corpora: 
\begin{inparaenum}[(1)]
\item training set of OpenIE6~\cite{kolluru2020openie6} and 
\item recently released LSOIE~\cite{lsoie-2021}.
\end{inparaenum}
For evaluation, following~\citet{kolluru2020openie6}, we employ 
the CaRB test set~\cite{bhardwaj-etal-2019-carb} together with OIE2016~\cite{Stanovsky2016EMNLP}, WiRe57~\cite{lechelle2019wire57}, CaRB and CaRB (1-1) evaluation scorers.
Our main contributions are as follows:
\begin{inparaenum}[(1)]
    \item we introduce DetIE, a novel approach for OpenIE that demonstrates improvements on common English benchmarks in both quality metrics and inference time; DetIE \added{does not replace the entire OIE pipeline} and can be combined with existing techniques such as grid constraints or coordination analysis, \added{which, as we show, might}  further improve the results;
    \item we investigate the language transferability of our model to other languages and obtain significant improvements on multilingual benchmarks;
    \item we propose a strategy for generating multilingual synthetic data and fine-tuning the model for a specific language.
\end{inparaenum}

\section{Related Work}\label{sec:relatedwork}
Early open information extraction approaches such as TextRunner~\cite{etzioni2008open},  ReVerb~\cite{fader2011identifying}, OLLIE~\cite{schmitz2012open},  ClausIE~\cite{del2013clausie}, and MinIE~\cite{gashteovski2017minie} were mostly rule-based, used automatically generated training data and separated modules such as PoS taggers, dependency parsers, and chunkers. While they have advantages such as domain independence, errors from separate modules tend to accumulate.

Modern approaches are usually based on neural networks with either recurrent~\cite{stanovsky-etal-2018-supervised,cui-etal-2018-neural} or, more recently, Transformer-based architectures~\cite{kolluru2020openie6,kolluru2020imojie}. Neural models can be trained end to end but require labeled data, and manual relation annotation for supervised learning is extremely costly. Therefore, neural open IE partly relies on classical approaches. 

\textit{Sequence labeling} approaches assume a triplet to be a subset of the input sequence, either predicting spans for all three parts in the input sequence or assigning a corresponding label---subject, relation, object, or background---to every token and assembling a triplet from these labels. 
These models cannot change the sentence structure or introduce new auxiliary words while generating predictions.

The \emph{RnnOIE} model~\cite{stanovsky-etal-2018-supervised} predicts entities given ground truth predicates during learning, but predicates are extracted with a PoS tagger during inference. In order for the model to be able to extract multiple overlapping tuples for each sentence, the authors used an extended version of \emph{BIO tagging} (beginning-inside-outside)~\cite{ramshaw1999text}.
Since then, several models have extended and improved over \emph{RnnOIE}.
In order to alleviate the lack of labeled training data, \emph{SenseOIE}~\cite{roy-etal-2019-supervising} augments model inputs with extractions from existing IE systems such as word embedding, part-of-speech tags, syntactic role labels, and dependency structure.
\emph{Multi\textsuperscript{2}OIE}~\cite{ro2020multi} is a two-step procedure that first predicts the predicates and then the corresponding entities. This model is able to make multilingual predictions by using multilingual BERT embeddings even if it had been trained only on an English dataset. \emph{SpanOIE}~\cite{zhan2020span} is also a two-step model, but unlike sequence labeling models shown above it predicts a span instead of a BIO tag for every token.

In natural language texts, predicates are often present only implicitly. To solve this, finding predicates can be viewed as a classification task~\cite{zeng2014relation}, but this approach is unsuitable for an open vocabulary setting. There, researchers use \textit{sequence generation} approaches: encoder-decoder frameworks that produce triplets as sequences, typically split via special tokens.
The first such model was \emph{NeuralOIE}~\cite{cui-etal-2018-neural}, later improved in \emph{IMoJIE}~\cite{kolluru2020imojie}. In IMoJIE, the next extraction is conditioned on all previously extracted tuples, which leads to more diverse tuples. Sequence generation models are heavy and have relatively low performance in both learning and inference due to autoregressive output generation. This problem was partially resolved in the \emph{IGL-OIE} model~\cite{kolluru2020openie6}, where the next extraction is still conditioned on all previous extractions, but the tuples themselves are extracted in the sequence labeling fashion.

A key advantage of sequence labeling over sequence generation is that the problem is formalized as token classification, so all classification-related techniques can be applied.
On the other hand, it is hard to define similarity metrics between generated text and the ground truth; existing metrics do not correlate well with human judgement~\cite{mathur2020tangled,lukasik2020semantic}, so the sequence generation approach is inherently biased.

Autoregressive generation of triplets inherent in previous methods forces the model to predict triplets in a predefined order, leading to additional arbitrary penalties.
In this work, we alleviate this problem and propose an approach that is entirely novel compared to the works discussed above.

\section{Method}\label{sec:method}
\def\ntok{n_{\mathrm{tokens}}}
\def\ntrip{n_{\mathrm{triplets}}}
\def\mtrip{m_{\mathrm{triplets}}}
\def\nclass{n_{\mathrm{classes}}}

We follow the sequence labeling paradigm.
Given an input sequence $\{x_1 \dots x_T\}$, the goal is to predict a set $S$ of token masks $\{\{L_{1,1} \dots L_{T,1}\} \dots \{L_{1,N} \dots L_{T,N}\}\}, \ |S|=N$, labeling each token in a mask with exactly one of $C = 4$ classes: ``Background'', ``Subject'', ``Relation'', or ``Object''. If a mask contains non-background tokens, it produces a triplet; for some applications, it is needed to ensure that subject, relation, and object tokens are all present in the mask.

Our method has two main components: a feedforward neural architecture and an order-agnostic loss function. It follows the general idea of convolutional architectures used for object detection in computer vision: it
\begin{inparaenum}[(1)]
    \item makes mutually-aware predictions in a single pass and
    \item matches them with the ground truth via intersection-based matching during training.
\end{inparaenum}
The latter both encourages relevant predictions to be closer to the ground truth and serves to discard irrelevant predictions and duplicates.
This type of architectures has been used in one-stage object detection, especially in the family of \emph{single-shot detectors} (SSD)~\cite{SSD} and their later developments with feature pyramids, e.g., RetinaNet~\cite{lin2017focal}. In computer vision, SSD uses a set of predefined \emph{anchor boxes} that represent default bounding box predictions for each position in the feature map; the same network predicts both class labels and refined positions for each anchor box, usually on several different scales, and the network is trained in an end-to-end fashion with a single loss function. In DetIE, the counterparts of anchor boxes are masks representing \emph{possible triplets}.

\begin{figure}[!t]\centering
\includegraphics[width=0.99\linewidth]{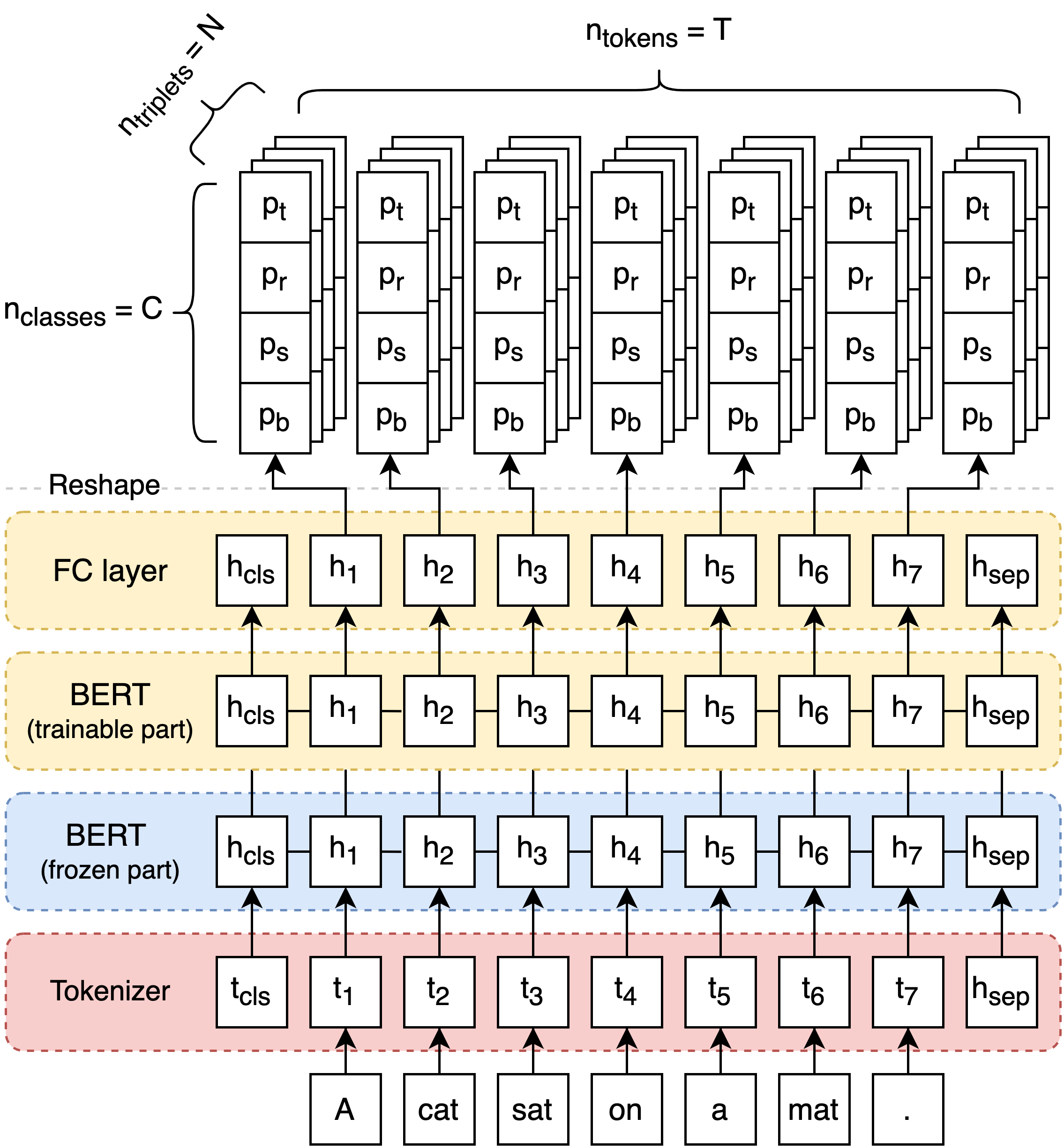}
\caption{Model architecture.}\label{figure:model}

\includegraphics[width=.9\linewidth]{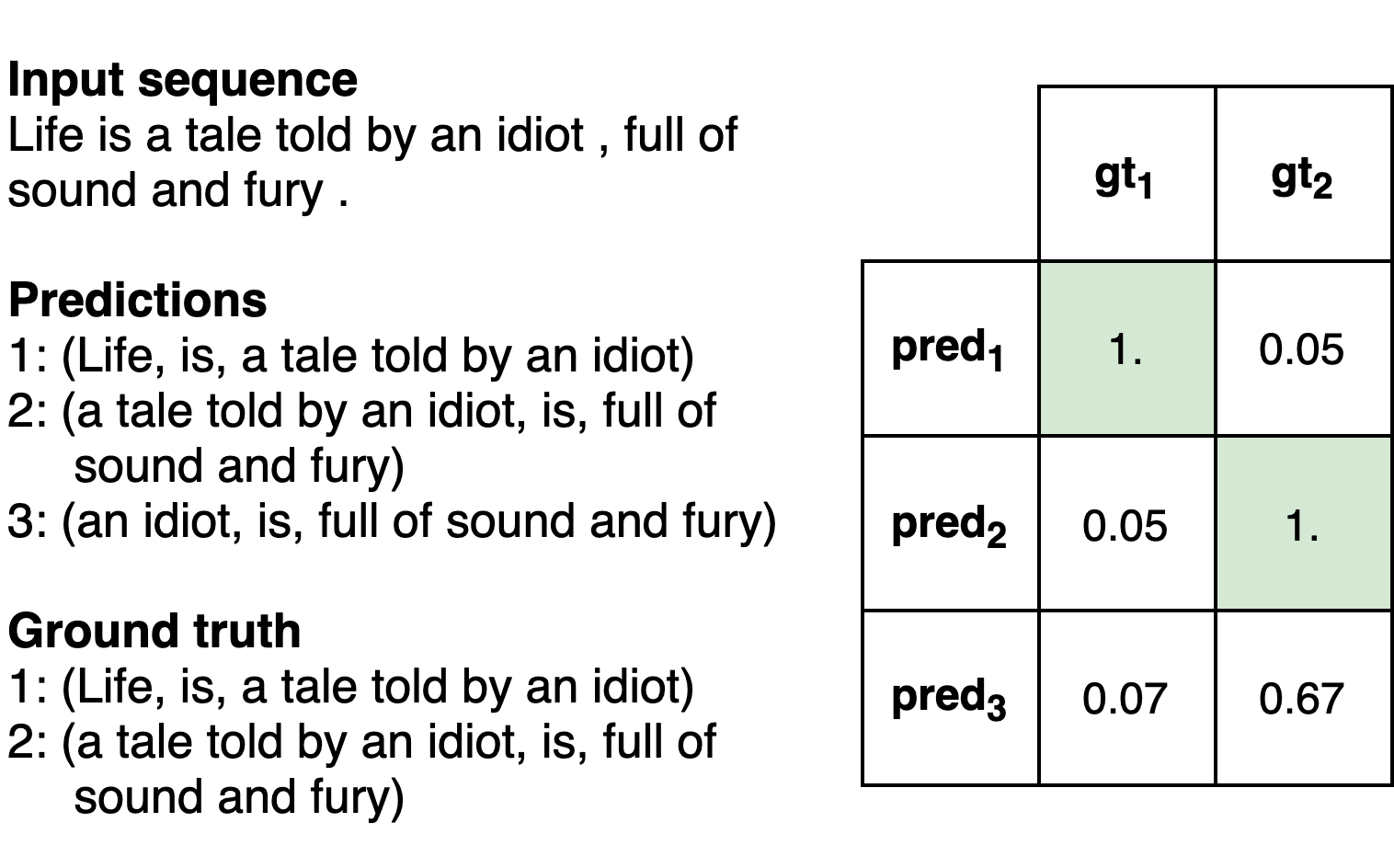}
\caption{Sample matching: predictions vs. ground truth; pred\textsubscript{3} is not matched and will be penalized. 
}\label{fig:bipartite}
\end{figure}

\subsection{Model}
Our model aims to extract a large predefined number of probability masks from a single text fragment, each corresponding to a possible triplet. For a given sequence of tokens, the model produces a three-dimensional tensor of probabilities $p$ (Fig.~\ref{figure:model}) of shape $(T, N, C)$, where $T$ corresponds to the number of tokens, $N$ is a pre-defined number of possible extracted triplets, and $C$ is the number of classes designated above. Since $N$ serves as an upper bound on the number of predictions, it should be large enough to cover all possible triplets in a document. However, larger values of $N$ exacerbate the class imbalance discussed in the next section. We found $N = 20$ to be sufficient for our datasets.

The output at position $(t, n, c)$ is the probability of a token $t$ to belong to class $c$ in mask $n$. The actual prediction is obtained from the probabilities by taking the $\argmax$ over classes for each token and filtering out background-only masks (i.e., if the maximal mask contains at least one non-background token, we extract it as a triplet).

The core of the architecture is a pre-trained BERT encoder as implemented in the \emph{HuggingFace} library~\cite{devlin-etal-2019-bert,wolf-etal-2020-transformers}.
We map its output to $N \times C$ channels for every input token with a fully-connected layer, and then reshape the obtained tensor to the shape of $p$ from above.
The value of $N$ does not affect the performance significantly since it only scales the last layer in our model linearly, leaving the most computationally intensive part---the BERT encoder---intact..
In the BERT architecture, we unfreeze several top layers to capture inter-token dependencies relevant for OpenIE. However, we provide an alternative version of our model with fully frozen BERT and an additional Transformer on top. This model has slightly worse performance but may be beneficial in the multilingual setting since it cannot lose multilingual information by unfreezing BERT.

\subsection{Order-Agnostic Loss}

To solve the issue of the predefined ordering of triplets, we design a special loss function with a bijective match between each predicted mask and the nearest ground truth triplet. Let $N$ and $M$ be the number of masks and ground truth triplets respectively; we choose $M$ most relevant predictions and encourage them to match the ground truth exactly with the cross-entropy loss. To find the best matching, we calculate the $N \times M$ IoU (intersection-over-union) matrix between each probability mask and one-hot representation of the ground truth. We then maximize 
the sum of IoUs over all matches with the Hungarian algorithm~\cite{kuhn1955hungarian}; see Fig.~\ref{fig:bipartite} for an example.
During training, we do not compute exact labels predicted by our model to avoid thresholding, but rather directly calculate the average smooth IoU between predicted probability masks and labels as follows:
$\mathrm{IoU}_{nm} = \frac{I_{nm}}{U_{nm}}$, where
    $I_{nm} = \sum_{t,c} p_{tnc} l_{tmc}$;
    $U_{nm} = \sum_{t,c}p_{tnc} + \sum_{t,c} l_{tmc} - I_{nm}$,
$p$ is the probability tensor predicted by the model, and $l$ is the one-hot tensor of ground truth labels.
The main drawback of this approach is that as the number of relations increases, the proportion of background (non-relation) tokens rapidly grows as well. We have taken measures against this induced class imbalance, discarding the background class in the IoU and reweighting non-background classes. We have also experimented with the focal loss instead of cross-entropy, but it did not yield any improvements, only shifted the precision-recall tradeoff.

\section{Setup and Datasets}\label{sec:experimental_setup}
\subsection{Experimental Setup}

We implement our model in pytorch\_lightning~\cite{falcon2019pytorch} with Hydra configuration framework~\cite{Yadan2019Hydra}. We use the pretrained \emph{bert-base-multilingual-cased} BERT by \emph{HuggingFace}~\cite{wolf-etal-2020-transformers} for both English and multilingual benchmarks.
During training, performance is measured as token-wise macro F1-score between predicted masks and ground truth labels. The best checkpoint is monitored along epochs. In case of IMoJIE data, 10\% of the samples are selected as validation. In case of LSOIE, the validation was performed on the test split of the dataset.
Our best model was trained with Adam optimizer with learning rate 5e-4 and weight decay 1e-6, batch size $32$, unfreezing $4$ top layers of BERT, $N=20$ detections, matching based on IoU similarity metrics, 
and doubled weights of non-background classes.
 Typical training time until the best model is reached is about $1.5$ hours on an NVIDIA Tesla V100 GPU.
Inference speed is measured in sentences processed per second on a well-known set of $3{,}200$ sentences~\cite{stanovsky-etal-2018-supervised}. We refer to results by~\citet{kolluru2020openie6} and assess the performance (Table~\ref{tab:carb}) in a similar setting: batches of $32$ processed on a single NVIDIA Tesla V100 GPU and 6 cores Intel(R) Xeon(R) CPU E5-2690 v4 @ 2.60GHz without any additional model optimizations.

\subsection{Datasets}\label{ssec:data}

Dataset statistics are summarized in Table~\ref{tab:datasets}.
First, we use the recent \textbf{LSOIE} (\emph{Large-Scale Open Information Extraction}) dataset~\cite{lsoie-2021}.
For a fair comparison, we have also trained models on the dataset used by~\citet{kolluru2020openie6,kolluru2020imojie}, called \textbf{IMoJIE} below. 
Another reason for training on two different datasets is that LSOIE
differs in its annotation scheme from 
popular evaluation datasets such as OIE2016~\cite{Stanovsky2016EMNLP} and CaRB~\cite{bhardwaj-etal-2019-carb};
\camera{e.g.}, in LSOIE auxiliary verbs such as ``is'' or ``was'' are (intentionally) not included into predicates, while CaRB adds more context into predicates than other OIE systems.
We also experimented with adding \emph{Wikidata}-based synthetic sentences during training (see below).
Thus, in our experiments we use crowdsourced (LSOIE), prediction-based (IMoJIE), and synthetic data. All these approaches are discussed in detail below.

\begin{table}[!t]
\centering\small
\begin{tabular*}{\columnwidth}{cp{2.4cm}cc} 
\toprule
\textbf{Split}         & \textbf{Dataset} & \textbf{\# sentences} & \textbf{\# tuples} \\ \hline
\multirow{3}{*}{Train} & IMoJIE  & $91{,}725$ & $190{,}661$  \\ 
                       & LSOIE & $34{,}780$ &  $100{,}862$ \\
                       & Synth & $10{,}000$ & $41{,}645$ \\
\hline
\multirow{4}{*}{Test}  & LSOIE & $7{,}900$  & $17{,}459$ \\                      
                       & CaRB   & $641$ & $2{,}715$ \\
                       & MultiOIE2016 & $595$ & $1{,}508$ \\
\bottomrule
\end{tabular*}

\caption{Dataset statistics; the MultiOIE2016 and Synth numbers are given for each language.}\label{tab:datasets}\vspace{.2cm}

    \begin{tabular}{r|l}\toprule
        \textbf{Prob} & \textbf{Action} \\ \hline
        0.1 & Concatenating the triplet and adding '.' \\
        0.2 & Triplet + conjunction + triplet. \\
        0.35 & Joining 3--5 concatenated triplets with ',' \\
        0.35 & Joining 2--9 concatenated triplets with '.'\\ \bottomrule
    \end{tabular}
    
    \caption{Templates and their probabilities for \emph{Synth}.}\label{tab:templates_wdt_synthetic}
\end{table}

\textbf{Wikidata-based synthetic sentences.} For this dataset, called \textbf{Synth} below,
we have used \emph{Wikidata}~\cite{vrandevcic2014wikidata} to devise a simple sentence generation strategy.
We lexicalize \emph{Wikidata} triplets, joining the subject, object, and relation (predicate) by white spaces, e.g., \texttt{(Albert Einstein, is, physicist)} becomes ``Albert Einstein is physicist''. These phrases are used to generate sentences, e.g., ``Albert Einstein is physicist while Amelia Mary Earhart is pilot.'' These sentences obviously have poor and highly standardized grammatical structure, but we have found that \camera{extra data of} this quality leads to better relation extraction on multilingual benchmarks (see below).
Sentences for \emph{Synth} are generated with a set of templates \camera{selected}
with fixed probabilities (Table~\ref{tab:templates_wdt_synthetic}).
\emph{Wikidata} is a very rich data source, with nearly $9{,}000$ different properties (predicates) that are highly imbalanced in the data, so \emph{Synth} is also imbalanced w.r.t. predicates.
 
We used \emph{Stanza}~\cite{qi2020stanza} for tokenization and PoS tagging. Tags are used for filtering aimed to create more grammatical sentences. Since synthetic sentences are simplistic, our model does not need too many examples of each type, although it important to have typological diversity in the samples. We generate data in two languages, Spanish and Portuguese.

\textbf{LSOIE.}
We have used the \camera{LSOIE} dataset~\cite{lsoie-2021} prepared based on QA-SRL 2.0~\cite{fitzgerald2018large} that expands the scope of OIE2016 and AW-OIE~\cite{stanovsky-etal-2018-supervised}.
The dataset can be converted to several different formats\footnote{\url{https://github.com/Jacobsolawetz/large-scale-oie}}; we have concatenated the \textbf{ lsoie\_(wiki$\vert$science)\_*.conll} subsets as sources for our training and test data. Most datapoints in LSOIE are single sentences with a tuple of extractable subsequences; relations in LSOIE are $N$-ary. One of the elements usually represents a predicate ({\tt P-B},~{\tt P-I}), others are arguments ({\tt A[N]-B},~{\tt A[N]-I}). 
Since we are interested only in subject-relation-object triplets, we have removed the datapoints that do not have a predicate and at least two arguments. Then, predicates
were converted into ``relation'' (``rel''/``pred''), arguments 
{\tt A0-*} 
were converted into ``subject'' (``source'', ``arg1''), and  arguments {\tt AN-*} (for $N>0$) were combined
into a single argument defined as ``object'' (``target'', ``arg2''). 

\textbf{IMoJIE (dataset).}
To provide a fair comparison with state of the art models such as OpenIE6~\cite{kolluru2020openie6} and~IMoJIE~\cite{kolluru2020imojie}, we also used the large-scale dataset they were trained on\footnote{\url{https://github.com/dair-iitd/imojie}} for training DetIE. The dataset consists of tuples extracted from \emph{Wikipedia} sentences via OpenIE4~\cite{christensen2011analysis}, ClausIE~\cite{del2013clausie}, and RnnOIE~\cite{stanovsky-etal-2018-supervised} and~filtered with~a~\textit{Score-and-Filter} technique proposed by~\citet{kolluru2020imojie}.
IMoJIE data is labelled for sequence \textit{generation}: each sentence is assigned with a set of string tuples representing triplets. We find that the triplets are typically
a combination of different pieces of the input sentence, so we apply a heuristic algorithm to retrieve them as masks for sequence labeling.
The algorithm iteratively finds longest common substrings of a sentence and a triplet, converts them into labeled spans and excludes, until either sentence or triplet are exhausted; since spans are excluded, (S,~R,~O) masks are guaranteed to be disjoint. 
This method fails only if some tokens (e.g., ``is'') are understood implicitly and do not occur in the sequence.
Similarly to \citet{kolluru2020openie6}, to cover such cases we append tokens ``[is]'', ``[from]'', and ``[to]'' to the end of a sequence, both in the converted IMoJIE and during prediction.

Next we describe our evaluation datasets: CaRB, LSOIE, and MultiOIE2016. We do not use two popular datasets, OIE2016 and WiRe57; the flaws of OIE2016 were discussed in~\cite{zhan2020span,bhardwaj-etal-2019-carb}, while WiRe57 is almost 2x smaller than CaRB test set: only 57 sentences with 343 manually extracted facts.

\textbf{CaRB (test).} This evaluation dataset\footnote{\url{https://github.com/dair-iitd/CaRB}} is a subset of sentences from the OIE2016 dataset~\cite{Stanovsky2016EMNLP} re-annotated via \emph{Amazon Mechanical Turk} crowdsourcing.  Annotators selected (arg1; rel; arg2) triplets and annotated time and location attributes if possible
\cite{bhardwaj-etal-2019-carb}.

\textbf{LSOIE (test).} We have used a combination of test sets of LSOIE$_{\mathrm{wiki}}$ and LSOIE$_{\mathrm{science}}$.

\textbf{MultiOIE2016 (test).} This dataset is based on Re-OIE2016~\cite{zhan2020span}, a version of OIE2016 tailored for sequence tagging (unlike generation-oriented CaRB). \citet{ro2020multi} extended it to Spanish and Portuguese;
the number of sentences and tuples is the same for each language.

\subsection{Evaluation Metrics}\label{ssec:eval_data}

Following~\citet{kolluru2020openie6}, we score model predictions 
against CaRB reference extractions,
evaluated using the schemes introduced in OIE2016, WiRe57, CaRB, and OpenIE6 and discussed in detail below.
Note that DetIE predicts a set of probabilities per token rather than confidence scores, but probabilities can be converted to ``confidence'' in many ways, e.g. aggregated max probability per argument or average $\log$ of probabilities.

\textbf{OIE2016.}
We use a scheme proposed for evaluation by \citet{Stanovsky2016EMNLP}\footnote{\url{https://github.com/gabrielStanovsky/oie-benchmark}}, who compare systems in terms of precision and recall; the crucial step is to match predicted and ground truth extractions. A prediction is matched with the ground truth if they agree on the grammatical head of the elements (arguments and predicate).  

\textbf{WiRe57.} \citet{lechelle2019wire57} introduced a new scoring procedure\footnote{\url{https://github.com/rali-udem/WiRe57}}
that penalizes overly long extractions and assigns a token-level prediction quality score to all gold-prediction pairs for each sentence. Unlike the OIE2016 scorer, it considers all pairs of extractions.
First, a predicted tuple is considered \textit{possibly matching} a reference tuple from the same sentence if they share at least one token from each relation, argument 0 and argument 1 (with triplets, as in our case, this means at least one token in common for every corresponding element of the tuple). Then precision, recall, and F1 scores are computed for all possibly matching pairs of predicted $t$ and reference $g$ tuples as follows:
$$\begin{array}{rcl}
    \mathrm{prec}(t, g) &=& \frac{1}{|t|}\sum_k{|t^{(k)} \cap g^{(k)}|},\\
    \mathrm{rec}(t, g) &=& \frac{1}{|g|}{\sum_k{|t^{(k)} \cap g^{(k)}|}},\\
    \mathrm{F1}(t, g) &=& \frac{2\mathrm{prec}(t, g)\mathrm{rec}(t, g)}{\mathrm{prec}(t, g)+\mathrm{rec}(t, g)},
\end{array}$$
where $t^{(k)}$ is the bag of words/tokens representation of the $k$th part of the tuple and $|t|$ and $|g|$ are the numbers of tokens in the corresponding tuples.

Having computed the scores for all possibly matching pairs, we greedily construct the best matching. Overall system performance is measured by micro-averaged precision and recall. For more details we refer to~\cite{lechelle2019wire57} and the original implementation\footnote{\url{https://github.com/rali-udem/WiRe57/blob/master/code/wire_scorer.py}}.

\textbf{CaRB.} This is a crowdsourced OIE2016 re-annotation initiative~\cite{bhardwaj-etal-2019-carb} already mentioned above; we use the evaluation scheme which here is different from the predecessors. CaRB uses filtering of stopwords and makes use of binary model predictions. 
CaRB scorer is not greedy; it constructs a matching table for all gold-predicted pairs, computes and averages maximum overall recall in each row, thus matching gold tuples with the best extraction. For precision, extractions are matched with gold annotations in a 1-1 fashion, then averaged as well. 
CaRB scorer uses only matches with at least one common word in the relation field. All higher order arguments (beyond (arg1;~rel;~arg2) triplets) are appended to the last argument. Illustrations of this single-to-many approach and a more detailed discussion are given in~\cite{bhardwaj-etal-2019-carb} and reference code\footnote{\url{https://github.com/dair-iitd/CaRB}}.

\textbf{CaRB (1-1).} Used by~\citet{kolluru2020openie6}, this evaluation scheme retains CaRB's similarity computation but uses a one-to-one mapping for both precision and recall, similar to OIE16 and Wire57.

\begin{table*}[t]
    \centering\small
    \setlength{\tabcolsep}{7pt}
    \begin{tabular}{l|cc|cc|cc|c|r} \toprule
        \multirow{3}{*}{Model} & \multicolumn{7}{|c|}{CaRB evaluation schemes} & Speed \\
        \cline{2-8} 
        & \multicolumn{2}{c|}{CaRB} & \multicolumn{2}{c|}{CaRB(1-1)} & \multicolumn{2}{c|}{OIE16-C} & \multicolumn{1}{c|}{Wire57-C} & \multirow{2}{*}{(sent./sec)} \\
        \cline{2-8}
           & {F1} & {AUC} & {F1} & {AUC} & {F1} & {AUC} & {F1} &  \\ \hline
        MinIE~\cite{gashteovski2017minie}  & 41.9 & - & 38.4 & - & 52.3 & - & 28.5 & 8.9 \\
        ClausIE~\cite{del2013clausie}  & 45.0 & 22.0 & 40.2 & 17.7 & 61.0 & 38.0 & 33.2 & 4.0 \\
        OpenIE4~\cite{christensen2011analysis}  & 51.6 & 29.5 & 40.5 & 20.1 & 54.3 & 37.1 & 34.4 & 20.1 \\
        OpenIE5~\cite{saha2017bootstrapping,saha2018open}  & 48.0 & 25.0 & 42.7 & 20.6 & 59.9 & 39.9 & 35.4 & 3.1\\ \hline
    SenseOIE~\cite{roy-etal-2019-supervising}  & 28.2 & - & 23.9 & - & 31.1 & - & 10.7 & - \\
        SpanOIE~\cite{zhan2020span}  & 48.5 & - & 37.9 & - & 54.0 & - & 31.9 & 19.4 \\
        RnnOIE~\cite{stanovsky-etal-2018-supervised}  & 49.0 & 26.0 & 39.5 & 18.3 & 56.0 & 32.0 & 26.4 & \underline{149.2} \\
        NeuralOIE~\cite{cui-etal-2018-neural}  & 51.6 & 32.8 & 38.7 & 19.8 & 53.5 & 37.0 & 33.3 & 11.5\\ \hline
        IMoJIE~\cite{kolluru2020imojie}  & \underline{53.5} & 33.3 & 41.4 & 22.2 & 56.8 & 39.6 & 36.0 & 2.6 \\
        IGL-OIE~\cite{kolluru2020openie6} & 52.4 & 33.7 & 41.1 & 22.9 & 55.0 & 36.0 & 34.9 & 142.0 \\
        CIGL-OIE~\cite{kolluru2020openie6} & \textbf{54.0} & \underline{35.7} & 42.8 & 24.6 & 59.2 & 40.0 & 36.8 & 142.0 \\
        {OpenIE6}~\cite{kolluru2020openie6} & 52.7 & 33.7 & \textbf{46.4} & \underline{26.8} & \underline{65.6} & \underline{48.4} & \textbf{40.0} & 31.7 \\ \hline 
        DetIE$_{\mathrm{LSOIE}}$ (ours) & 43.0 & 27.2\textsuperscript{$\ast$} & 33.1 & 18.3\textsuperscript{$\ast$} & 49.7 & 32.7\textsuperscript{$\ast$} & 31.2 & \textbf{708.6} \\ 
        DetIE$_{\mathrm{IMoJIE}}$  (ours)& 52.1 & \textbf{36.7}\textsuperscript{$\ast$} & 40.1 & 24.0\textsuperscript{$\ast$} & 56.0 & 38.7\textsuperscript{$\ast$} & 36.0 & \textbf{708.6}  \\ 
        DetIE$_{\mathrm{LSOIE}}$  (ours) + IGL-CA from OpenIE6 &  39.6 & 26.7\textsuperscript{$\ast$} & 36.3 & 22.7\textsuperscript{$\ast$} & 63.3 & 47.9\textsuperscript{$\ast$} & 33.5 &  112.2\\ 
        DetIE$_{\mathrm{IMoJIE}}$  (ours) + IGL-CA from OpenIE6 & 47.3 & 35.1\textsuperscript{$\ast$} & \underline{43.1} & \textbf{29.3}\textsuperscript{$\ast$} & \textbf{67.7} & \textbf{54.0}\textsuperscript{$\ast$} & \underline{37.8} & 112.2 \\ 
        \bottomrule
    \end{tabular}
    \caption{Comparison on CaRB test set with scoring schemes from CaRB~\cite{bhardwaj-etal-2019-carb}, CaRB (1-1)~\cite{kolluru2020openie6}, OIE2016~\cite{Stanovsky2016EMNLP}, WiRe57~\cite{lechelle2019wire57}. Results for all models except DetIE are cited from~\cite{kolluru2020openie6}. 
    Best results are shown in bold; second best, underlined. DetIE does not provide confidence scores, so ROC-AUC values are approximated from a single TPR-FPR point.}
    \label{tab:carb}
\end{table*}

\begin{table}[!t]
    \centering\small
    \setlength{\tabcolsep}{7pt}
    \begin{tabular}{l|ll} \toprule
         Model & {F1} & {AUC} \\ \hline
        OllIE~\cite{ollie-emnlp12} & 36.8 & 16.7 \\
        ReVerb~\cite{fader2011identifying} & 36.8 & 16.9 \\
        OpenIE4 & 54.6  & 32.3 \\
        OpenIE5 & 49.5  & 25.8 \\ 
        CIGL-OIE & \underline{59.7}  & 48.0 \\
        OpenIE6 (CIGL-OIE + IGL-CA) & 51.6 & 32.7\\ \hline 
        DetIE$_{\mathrm{IMoJIE}}$ & 55.7 & 44.9$^*$ \\
        DetIE$_{\mathrm{IMoJIE}}$  (ours)  + IGL-CA & 45.9 & 41.7$^*$ \\ 
        DetIE$_{\mathrm{LSOIE}}$ & \textbf{71.4} & \textbf{61.3}$^*$ \\ 
        DetIE$_{\mathrm{LSOIE}}$ + IGL-CA & 58.7 & \underline{55.9}$^*$ \\ 
        \bottomrule
    \end{tabular}
    \caption{Comparison on combined LSOIE test sets~\cite{lsoie-2021} with the original CaRB evaluation scheme~\cite{bhardwaj-etal-2019-carb}. 
    Best results are shown in bold; second best, underlined. \added{DetIE does not provide confidence scores, so ROC-AUC values are approximated from a single TPR-FPR point.}}
    \label{tab:lsoie_table}
\end{table}

\begin{table}[!t]
\centering\small
\begin{tabularx}{\columnwidth}{c|l|XXX}
\toprule
Lang.               & \multicolumn{1}{c|}{Model}
                    & F1 & Prec. & Rec. \\ \hline
    \multirow{5}{*}{EN} & \small ArgOE
                    & 43.4 & 56.6 & 35.2                            \\
                    & \small PredPatt
                    & 53.1 & 53.9 & 52.3                            \\
                    & \small Multi$^2$OIE
                    & 69.3 & 66.9 & \underline{71.7} \\ 
                    & \small DetIE$_\mathrm{IMoJIE}$ (ours)
                    & \underline{78.7} & \underline{85.4} & 69.9 \\
                    & \small DetIE$_\mathrm{IMoJIE+Synth}$ (ours)
                    & \textbf{79.3} & \textbf{87.1} & \textbf{72.8} \\
                    \hline
    \multirow{5}{*}{ES} & \small ArgOE  
                    & 39.4 & 48.0 & 33.4                            \\
                    & \small PredPatt  
                    & 44.3 & 44.8 & 43.8                            \\
                    & \small Multi$^2$OIE 
                    & 60.2 & 59.1 & 61.2 \\ 
                    & \small DetIE$_\mathrm{IMoJIE}$ (ours) 
                    & \underline{73.2} & \underline{83.7} & \underline{65.0} \\
                    & \small DetIE$_\mathrm{IMoJIE+Synth}$ (ours)
                    & \textbf{75.0} & \textbf{85.6} & \textbf{66.8} \\
                    \hline
    \multirow{5}{*}{PT} & \small ArgOE    
                    & 38.3 & 46.3 & 32.7                   \\
                    & \small PredPatt  
                    & 42.9 & 43.6 & 42.3                            \\
                    & \small Multi$^2$OIE
                    & 59.1 & 56.1 & 62.5          \\
                    & \small DetIE$_\mathrm{IMoJIE}$ (ours)
                    & \underline{74.7} & \underline{85.1} & \underline{66.6} \\
                    & \small DetIE$_\mathrm{IMoJIE+Synth}$ (ours)
                    & \textbf{75.0} & \textbf{86.0} & \textbf{69.4} \\
\bottomrule
\end{tabularx}

\caption{Binary extraction performance on MultiOIE2016~\cite{zhan2020span} \added{measured with CaRB’s evaluation scheme}. Results for models other than DetIE are cited from~\cite{ro2020multi}.}
\label{tab:multioie}
\end{table}

\section{Experiments and Results}\label{sec:experiments}

\subsection{Models and Systems in Comparison}\label{ssec:compared_systems}
We compare the proposed Det-IE model on the LSOIE and CaRB datasets with the following non-neural models: 
\begin{inparaenum}[(1)]
\item MinIE \cite{gashteovski2017minie}, \item  ClausIE \cite{del2013clausie}, \item  OpenIE4\footnote{\url{https://github.com/allenai/openie-standalone}} \cite{christensen2011analysis},
\item OpenIE5\footnote{\url{https://github.com/dair-iitd/OpenIE-standalone}} \cite{saha2017bootstrapping,saha2018open}; and the following neural models:
\item IMoJIE \cite{kolluru2020imojie},
\item NeuralOIE \cite{cui-etal-2018-neural},
\item RnnOIE \cite{stanovsky-etal-2018-supervised},
\item SenseOIE \cite{roy-etal-2019-supervising},
\item SpanOIE \cite{zhan2020span},
\item CIGL-OIE, 
\item OpenIE6 (CIGL-OIE + IGL-CA)\footnote{\url{https://github.com/dair-iitd/openie6}} \cite{kolluru2020openie6}.
\end{inparaenum}
SpanOIE is a span-based model. IMoJIE and NeuralOIE are generative models. RnnOIE, SenseOIE, CIGL-OIE, and OpenIE6 are sequence labeling models. 

\added{In experiments with monolingual data}, 
we additionally preprocess raw sentences in the test set using a coordination analysis (CA) model IGL-CA \added{following OpenIE6 approach}~\cite{kolluru2020openie6}\footnote{\url{https://github.com/dair-iitd/openie6}} which considers CA as a grid labeling problem and is trained on the coordination-annotated Penn Treebank~\cite{ficler2016coordination}. We first apply CA to sentences in the test set and then apply DetIE to the resulting ``simplified'' texts, attributing the extractions to the corresponding original sentences with neither post-filtering nor rescoring.

We also compare DetIE with three systems on MultiOIE2016: rule-based multilingual systems
\begin{inparaenum}[(1)]
\item ArgOE \cite{gamallo2015multilingual} and 
\item PredPatt \cite{white2016universal} and
\item neural BERT-based Multi$^2$OIE system~\cite{ro2020multi}.
\end{inparaenum}

\subsection{Results}

Tables~\ref{tab:carb} and \ref{tab:lsoie_table} report the quality metrics and performance comparisons across all metrics for the CaRB and LSOIE datasets respectively (DetIE and IGL-CA inference times were estimated separately, on an NVIDIA Tesla V100).
In Table~\ref{tab:carb}, IGL-CA is a pretrained coordination analysis model by~\citet{kolluru2020openie6}, used as discussed above. Table~\ref{tab:carb} shows that variations of DetIE improve upon the state of the art for almost all evaluation schemes on CaRB. For LSOIE,
as expected, DetIE$_{\mathrm{LSOIE}}$ performs best on LSOIE test set (Table~\ref{tab:lsoie_table}) since it was trained on data with the same annotation principles. Note, however, that even training on a different dataset (IMoJIE) allows DetIE models to be among the best-performing ones. \added{Our hypothesis is that our training scheme being able to capture multiple relations at once allows the underlying Transformer architecture to better use its ability to perceive relations in a text for extraction.}

Results for MultiOIE2016 are presented in Table~\ref{tab:multioie}. Here, the DetIE$_\mathrm{IMoJIE}$ model trained on IMoJIE significantly outperforms previous approaches. The margin in the F1 evaluation metric reaches 15.6\% on the Portuguese part of the dataset, while for English it is 8.6\%. Interestingly, Multi$^2$OIE shows better performance on English in terms of recall, although the gap between it and DetIE is only 1.8\%.

Effectively, training on IMoJIE makes DetIE a zero-shot model for Spanish and Portuguese, since IMoJIE is collected entirely in English. Thus, we decided to fuse training data with \emph{Synth} for our model; the result is shown as DetIE$_\mathrm{IMoJIE+Synth}$. Synthetic data adds another 1.8\% of F1 for Spanish, 0.6\% for English, and 0.3\% for Portuguese. This model outperforms previous state of the art w.r.t. all metrics, including recall on English.

Note that Multi$^2$OIE is a two-stage approach, thus it is run at least twice for each sentence: one time to extract predicates and possibly several times for the predicates, one for each. In contrast, DetIE is a single-shot model; we extract all relations at once, only capping at the number of possible extractions\footnote{It serves as a hyperparameter of our model; we used \camera{100} possible extractions as a large upper bound for a single sentence.}. \added{We believe that this is the main reason why our model is so much faster during inference.}

\begin{table}[!t]\footnotesize\setlength{\tabcolsep}{1pt}
\begin{tabular}{|p{.1\linewidth}|p{.87\linewidth}|}
\hline
\textbf{Sent. \#1} & Males had a median income of \$ 28,750 versus \$ 16,250 for females. \\ \hline
\multirow{2}{*}{Gold} & (Males; had a median income of; \$ 28,750) \\
 &  (females; had a median income of; \$ 16,250) \\  \hline
DetIE & (Males; had; a median income of \$ 28,750 versus \$ 16,250 for females) \\ \hline  \hline
\textbf{Sent. \#2} & Hapoel Lod played in the top division during the 1960s and 1980s, and won the State Cup in 1984. \\ \hline
\multirow{3}{*}{Gold} & (Hapoel Lod; played in; the top division; during the 1960s) \\
& (Hapoel Lod; played in; the top division; during the 1980s) \\
& (Hapoel Lod; won; the State Cup; in 1984) \\ \hline
\multirow{2}{*}{DetIE} & (Hapoel Lod; played; in the top division during the 1960s and 1980s) \\ 
&  (Hapoel Lod; won; the State Cup in 1984) \\ \hline  \hline
\textbf{Sent. \#3} & A spectrum from a single FID has a low signal-to-noise ratio, but fortunately it improves readily with averaging of repeated acquisitions. \\ \hline
\multirow{2}{*}{Gold} & (A spectrum from a single FID; has; a low signal-to-noise ratio) \\ 
& (signal-to-noise ratio; improves readily with averaging of; repeated acquisitions) \\  \hline
\multirow{2}{*}{DetIE} & (A spectrum from a single FID; has; a low signal-to-noise ratio) \\
 & (it; improves; readily with averaging of repeated acquisitions) \\ \hline \hline 
 
\end{tabular}
\caption{Sample sentences with gold annotations and relations predicted by DetIE$_{\mathrm{IMoJIE}}$.
}\label{tab:errors}
\end{table}

\subsection{Discussion and Error Analysis}\label{ssec:discussion}

We have analyzed the outputs of DetIE$_{\mathrm{IMoJIE}}$ on a random sample of $100$ sentences from the CaRB validation set (Table~\ref{tab:errors}). According to our analysis, DetIE is prone to aggregating conjunctions and comparisons into a single triplet (Sent.~\#1,~\#2), which explains the improvements from coordination analysis in Table~\ref{tab:carb}. Occasionally, we observed incorrect prediction of triplets in sentences with coreference (Sent.~\#3).
Since our model uses a single pass for relation extraction, we hypothesise that it could be applied to whole passages of text (multiple sentences) and extract relations which permeate the limits of a single sentence. Thus, the co-reference task could be done alongside with relation extraction in end-to-end manner; we leave this for further work.

The performance of DetIE on CaRB varies widely across training sets, which is expected since CaRB is only a test set and has its own markup scheme, different from schemes used in training datasets. There is no conventional gold standard training set for OpenIE: IMoJIE was obtained with an advanced bootstrapping scheme, OpenIE6 (SotA) was trained on it, and LSOIE is the latest published dataset of size suitable for neural models but it differs significantly, hence the difference in performance.

We hypothesise that one-hot-encoded PoS and/or dependencies head labels or appending the dependency head’s embedding to each token’s embedding could have improved the results. We believe so since many early OpenIE models did heavily rely on syntax (e.g, OpenIE5, OpenIE6), but it remains to be tested in further work.

\section{Conclusion}\label{sec:conclusion}

We have introduced a novel DetIE model for the OpenIE task; it is based on the ideas of single-shot object detection in computer vision and extracts multiple triplets in a single pass. The proposed model is atomic and can be used as a part of OIE pipelines that may include coordination analysis, rescoring, syntactic chunks collapsing etc. Our approach outperforms existing state of the art on the LSOIE dataset and performs at least on par or better for every considered evaluation scheme on CaRB.
The DetIE model is 5x faster than previous state of the art in terms of inference speed.

Moreover, DetIE has shown excellent performance in the zero-shot cross-lingual setting, exceeding existing state of the art for Spanish and Portuguese by 13\% and 15\% respectively. We have also introduced a technique for multilingual synthetic data generation and used it to generate additional training data that further improved the results (by 1.8\% in Spanish and 0.3\% in Portuguese).

As a first step for future work, our method may benefit from enrichment with PoS tags, syntactic information (e.g., {\tt deprel} tags), and traditional neural sequence labeling layers (e.g., CRF). We also plan to experiment with other possible improvements in the model architecture.

\section*{Acknowledgements}

The work of Sergey Nikolenko was supported by a grant for research centers in the field of artificial intelligence, provided by the Analytical Center for the Government of the Russian Federation in accordance with the subsidy agreement (agreement identifier 000000D730321P5Q0002) and the agreement with the Ivannikov Institute for System Programming of the Russian Academy of Sciences dated November 2, 2021 No. 70-2021-00142. The work of Elena Tutubalina was supported by a grant from the President of the Russian Federation for young scientists-candidates of science (MK-3193.2021.1.6). We are grateful to Daria Grischenko and Maxim Eremenko for their highly valuable help with organizing this work.

\FloatBarrier

\bibliography{aaai22}

\begin{thebibliography}{44}
\providecommand{\natexlab}[1]{#1}

\bibitem[{Balasubramanian et~al.(2013)Balasubramanian, Soderland, {Mausam}, and
  Etzioni}]{balasubramanian-etal-2013-generating}
Balasubramanian, N.; Soderland, S.; {Mausam}; and Etzioni, O. 2013.
\newblock Generating Coherent Event Schemas at Scale.
\newblock In \emph{Proceedings of the 2013 Conference on Empirical Methods in
  Natural Language Processing}, 1721--1731. Seattle, Washington, USA:
  Association for Computational Linguistics.

\bibitem[{Bhardwaj, Aggarwal, and Mausam(2019)}]{bhardwaj-etal-2019-carb}
Bhardwaj, S.; Aggarwal, S.; and Mausam, M. 2019.
\newblock {C}a{RB}: A Crowdsourced Benchmark for Open {IE}.
\newblock In \emph{Proceedings of the 2019 Conference on Empirical Methods in
  Natural Language Processing and the 9th International Joint Conference on
  Natural Language Processing (EMNLP-IJCNLP)}, 6263--6268. Hong Kong, China:
  Association for Computational Linguistics.

\bibitem[{Christensen, Soderland, and Etzioni(2011)}]{christensen2011analysis}
Christensen, J.; Soderland, S.; and Etzioni, O. 2011.
\newblock An analysis of open information extraction based on semantic role
  labeling.
\newblock In \emph{Proceedings of the sixth international conference on
  Knowledge capture}, 113--120.

\bibitem[{Cui, Wei, and Zhou(2018)}]{cui-etal-2018-neural}
Cui, L.; Wei, F.; and Zhou, M. 2018.
\newblock Neural Open Information Extraction.
\newblock In \emph{Proceedings of the 56th Annual Meeting of the Association
  for Computational Linguistics (Volume 2: Short Papers)}, 407--413. Melbourne,
  Australia: Association for Computational Linguistics.

\bibitem[{Del~Corro and Gemulla(2013)}]{del2013clausie}
Del~Corro, L.; and Gemulla, R. 2013.
\newblock Clausie: clause-based open information extraction.
\newblock In \emph{Proceedings of the 22nd international conference on World
  Wide Web}, 355--366.

\bibitem[{Devlin et~al.(2019)Devlin, Chang, Lee, and
  Toutanova}]{devlin-etal-2019-bert}
Devlin, J.; Chang, M.-W.; Lee, K.; and Toutanova, K. 2019.
\newblock {BERT}: Pre-training of Deep Bidirectional Transformers for Language
  Understanding.
\newblock In \emph{Proceedings of the 2019 Conference of the North {A}merican
  Chapter of the Association for Computational Linguistics: Human Language
  Technologies, Volume 1 (Long and Short Papers)}, 4171--4186. Minneapolis,
  Minnesota: Association for Computational Linguistics.

\bibitem[{Etzioni et~al.(2008)Etzioni, Banko, Soderland, and
  Weld}]{etzioni2008open}
Etzioni, O.; Banko, M.; Soderland, S.; and Weld, D.~S. 2008.
\newblock Open information extraction from the web.
\newblock \emph{Communications of the ACM}, 51(12): 68--74.

\bibitem[{Fader, Soderland, and Etzioni(2011)}]{fader2011identifying}
Fader, A.; Soderland, S.; and Etzioni, O. 2011.
\newblock Identifying relations for open information extraction.
\newblock In \emph{Proceedings of the 2011 conference on empirical methods in
  natural language processing}, 1535--1545.

\bibitem[{Falcon(2019)}]{falcon2019pytorch}
Falcon, W. A. e.~a. 2019.
\newblock PyTorch Lightning.
\newblock GitHub.

\bibitem[{Fan et~al.(2019)Fan, Gardent, Braud, and
  Bordes}]{DBLP:journals/corr/abs-1910-08435}
Fan, A.; Gardent, C.; Braud, C.; and Bordes, A. 2019.
\newblock Using Local Knowledge Graph Construction to Scale Seq2Seq Models to
  Multi-Document Inputs.
\newblock \emph{CoRR}, abs/1910.08435.

\bibitem[{Ficler and Goldberg(2016)}]{ficler2016coordination}
Ficler, J.; and Goldberg, Y. 2016.
\newblock Coordination Annotation Extension in the Penn Tree Bank.
\newblock In \emph{Proceedings of the 54th Annual Meeting of the Association
  for Computational Linguistics (Volume 1: Long Papers)}, 834--842.

\bibitem[{FitzGerald et~al.(2018)FitzGerald, Michael, He, and
  Zettlemoyer}]{fitzgerald2018large}
FitzGerald, N.; Michael, J.; He, L.; and Zettlemoyer, L. 2018.
\newblock Large-Scale QA-SRL Parsing.
\newblock In \emph{Proceedings of the 56th Annual Meeting of the Association
  for Computational Linguistics (Volume 1: Long Papers)}, 2051--2060.

\bibitem[{Gamallo and Garcia(2015)}]{gamallo2015multilingual}
Gamallo, P.; and Garcia, M. 2015.
\newblock Multilingual open information extraction.
\newblock In \emph{Portuguese Conference on Artificial Intelligence}, 711--722.
  Springer.

\bibitem[{Gashteovski, Gemulla, and Corro(2017)}]{gashteovski2017minie}
Gashteovski, K.; Gemulla, R.; and Corro, L.~d. 2017.
\newblock Minie: minimizing facts in open information extraction.
\newblock Association for Computational Linguistics.

\bibitem[{Kolluru et~al.(2020{\natexlab{a}})Kolluru, Adlakha, Aggarwal, Mausam,
  and Chakrabarti}]{kolluru2020openie6}
Kolluru, K.; Adlakha, V.; Aggarwal, S.; Mausam; and Chakrabarti, S.
  2020{\natexlab{a}}.
\newblock OpenIE6: Iterative Grid Labeling and Coordination Analysis for Open
  Information Extraction.
\newblock arXiv:2010.03147.

\bibitem[{Kolluru et~al.(2020{\natexlab{b}})Kolluru, Aggarwal, Rathore,
  Chakrabarti et~al.}]{kolluru2020imojie}
Kolluru, K.; Aggarwal, S.; Rathore, V.; Chakrabarti, S.; et~al.
  2020{\natexlab{b}}.
\newblock IMoJIE: Iterative Memory-Based Joint Open Information Extraction.
\newblock In \emph{Proceedings of the 58th Annual Meeting of the Association
  for Computational Linguistics}, 5871--5886.

\bibitem[{Kuhn(1955)}]{kuhn1955hungarian}
Kuhn, H.~W. 1955.
\newblock The Hungarian method for the assignment problem.
\newblock \emph{Naval research logistics quarterly}, 2(1-2): 83--97.

\bibitem[{Lechelle, Gotti, and Langlais(2019)}]{lechelle2019wire57}
Lechelle, W.; Gotti, F.; and Langlais, P. 2019.
\newblock WiRe57: A Fine-Grained Benchmark for Open Information Extraction.
\newblock In \emph{Proceedings of the 13th Linguistic Annotation Workshop},
  6--15.

\bibitem[{Lin et~al.(2017)Lin, Goyal, Girshick, He, and
  Doll{\'a}r}]{lin2017focal}
Lin, T.-Y.; Goyal, P.; Girshick, R.; He, K.; and Doll{\'a}r, P. 2017.
\newblock Focal loss for dense object detection.
\newblock In \emph{Proceedings of the IEEE international conference on computer
  vision}, 2980--2988.

\bibitem[{Liu et~al.(2015)Liu, Anguelov, Erhan, Szegedy, Reed, Fu, and
  Berg}]{SSD}
Liu, W.; Anguelov, D.; Erhan, D.; Szegedy, C.; Reed, S.~E.; Fu, C.; and Berg,
  A.~C. 2015.
\newblock {SSD:} Single Shot MultiBox Detector.
\newblock \emph{CoRR}, abs/1512.02325.

\bibitem[{Lukasik et~al.(2020)Lukasik, Jain, Menon, Kim, Bhojanapalli, Yu, and
  Kumar}]{lukasik2020semantic}
Lukasik, M.; Jain, H.; Menon, A.~K.; Kim, S.; Bhojanapalli, S.; Yu, F.; and
  Kumar, S. 2020.
\newblock Semantic Label Smoothing for Sequence to Sequence Problems.
\newblock arXiv:2010.07447.

\bibitem[{Mathur, Baldwin, and Cohn(2020)}]{mathur2020tangled}
Mathur, N.; Baldwin, T.; and Cohn, T. 2020.
\newblock Tangled up in BLEU: Reevaluating the Evaluation of Automatic Machine
  Translation Evaluation Metrics.
\newblock arXiv:2006.06264.

\bibitem[{Mausam(2016)}]{Mausam2016OpenIE}
Mausam. 2016.
\newblock Open Information Extraction Systems and Downstream Applications.
\newblock In \emph{IJCAI}.

\bibitem[{Mausam et~al.(2012)Mausam, Schmitz, Bart, Soderland, and
  Etzioni}]{ollie-emnlp12}
Mausam; Schmitz, M.; Bart, R.; Soderland, S.; and Etzioni, O. 2012.
\newblock Open Language Learning for Information Extraction.
\newblock In \emph{Proceedings of Conference on Empirical Methods in Natural
  Language Processing and Computational Natural Language Learning
  (EMNLP-CONLL)}.

\bibitem[{Ponza, Del~Corro, and Weikum(2018)}]{ponza-etal-2018-facts}
Ponza, M.; Del~Corro, L.; and Weikum, G. 2018.
\newblock Facts That Matter.
\newblock In \emph{Proceedings of the 2018 Conference on Empirical Methods in
  Natural Language Processing}, 1043--1048. Brussels, Belgium: Association for
  Computational Linguistics.

\bibitem[{Qi et~al.(2020)Qi, Zhang, Zhang, Bolton, and Manning}]{qi2020stanza}
Qi, P.; Zhang, Y.; Zhang, Y.; Bolton, J.; and Manning, C.~D. 2020.
\newblock Stanza: A {Python} Natural Language Processing Toolkit for Many Human
  Languages.
\newblock In \emph{Proceedings of the 58th Annual Meeting of the Association
  for Computational Linguistics: System Demonstrations}.

\bibitem[{Ramshaw and Marcus(1999)}]{ramshaw1999text}
Ramshaw, L.~A.; and Marcus, M.~P. 1999.
\newblock Text chunking using transformation-based learning.
\newblock In \emph{Natural language processing using very large corpora},
  157--176. Springer.

\bibitem[{Ro, Lee, and Kang(2020)}]{ro2020multi}
Ro, Y.; Lee, Y.; and Kang, P. 2020.
\newblock Multi\^{} 2OIE: Multilingual Open Information Extraction based on
  Multi-Head Attention with BERT.
\newblock \emph{arXiv preprint arXiv:2009.08128}.

\bibitem[{Roy et~al.(2019)Roy, Park, Lee, and Pan}]{roy-etal-2019-supervising}
Roy, A.; Park, Y.; Lee, T.; and Pan, S. 2019.
\newblock Supervising Unsupervised Open Information Extraction Models.
\newblock In \emph{Proceedings of the 2019 Conference on Empirical Methods in
  Natural Language Processing and the 9th International Joint Conference on
  Natural Language Processing (EMNLP-IJCNLP)}, 728--737. Hong Kong, China:
  Association for Computational Linguistics.

\bibitem[{Saha, Pal et~al.(2017)}]{saha2017bootstrapping}
Saha, S.; Pal, H.; et~al. 2017.
\newblock Bootstrapping for numerical open {IE}.
\newblock In \emph{Proceedings of the 55th Annual Meeting of the Association
  for Computational Linguistics (Volume 2: Short Papers)}, 317--323.

\bibitem[{Saha et~al.(2018)}]{saha2018open}
Saha, S.; et~al. 2018.
\newblock Open information extraction from conjunctive sentences.
\newblock In \emph{Proceedings of the 27th International Conference on
  Computational Linguistics}, 2288--2299.

\bibitem[{Schmitz et~al.(2012)Schmitz, Soderland, Bart, Etzioni
  et~al.}]{schmitz2012open}
Schmitz, M.; Soderland, S.; Bart, R.; Etzioni, O.; et~al. 2012.
\newblock Open language learning for information extraction.
\newblock In \emph{Proceedings of the 2012 joint conference on empirical
  methods in natural language processing and computational natural language
  learning}, 523--534.

\bibitem[{Solawetz and Larson(2021)}]{lsoie-2021}
Solawetz, J.; and Larson, S. 2021.
\newblock {LSOIE}: A Large-Scale Dataset for Supervised Open Information
  Extraction.
\newblock In \emph{Proceedings of the 16th Conference of the European Chapter
  of the Association for Computational Linguistics: Main Volume}, 2595--2600.
  Online: Association for Computational Linguistics.

\bibitem[{Stanovsky and Dagan(2016)}]{Stanovsky2016EMNLP}
Stanovsky, G.; and Dagan, I. 2016.
\newblock Creating a Large Benchmark for Open Information Extraction.
\newblock In \emph{Proceedings of the 2016 Conference on Empirical Methods in
  Natural Language Processing (EMNLP)}, (to appear). Austin, Texas: Association
  for Computational Linguistics.

\bibitem[{Stanovsky, Dagan, and {Mausam}(2015)}]{stanovsky-etal-2015-open}
Stanovsky, G.; Dagan, I.; and {Mausam}. 2015.
\newblock Open {IE} as an Intermediate Structure for Semantic Tasks.
\newblock In \emph{Proceedings of the 53rd Annual Meeting of the Association
  for Computational Linguistics and the 7th International Joint Conference on
  Natural Language Processing (Volume 2: Short Papers)}, 303--308. Beijing,
  China: Association for Computational Linguistics.

\bibitem[{Stanovsky et~al.(2018)Stanovsky, Michael, Zettlemoyer, and
  Dagan}]{stanovsky-etal-2018-supervised}
Stanovsky, G.; Michael, J.; Zettlemoyer, L.; and Dagan, I. 2018.
\newblock Supervised Open Information Extraction.
\newblock In \emph{Proceedings of the 2018 Conference of the North {A}merican
  Chapter of the Association for Computational Linguistics: Human Language
  Technologies, Volume 1 (Long Papers)}, 885--895. New Orleans, Louisiana:
  Association for Computational Linguistics.

\bibitem[{Tan, Pang, and Le(2020)}]{tan2020efficientdet}
Tan, M.; Pang, R.; and Le, Q.~V. 2020.
\newblock EfficientDet: Scalable and efficient object detection.
\newblock In \emph{Proceedings of the IEEE/CVF conference on computer vision
  and pattern recognition}, 10781--10790.

\bibitem[{Vrande{\v{c}}i{\'c} and Kr{\"o}tzsch(2014)}]{vrandevcic2014wikidata}
Vrande{\v{c}}i{\'c}, D.; and Kr{\"o}tzsch, M. 2014.
\newblock Wikidata: a free collaborative knowledgebase.
\newblock \emph{Communications of the ACM}, 57(10): 78--85.

\bibitem[{White et~al.(2016)White, Reisinger, Sakaguchi, Vieira, Zhang,
  Rudinger, Rawlins, and Van~Durme}]{white2016universal}
White, A.~S.; Reisinger, D.; Sakaguchi, K.; Vieira, T.; Zhang, S.; Rudinger,
  R.; Rawlins, K.; and Van~Durme, B. 2016.
\newblock Universal decompositional semantics on universal dependencies.
\newblock In \emph{Proceedings of the 2016 Conference on Empirical Methods in
  Natural Language Processing}, 1713--1723.

\bibitem[{Wolf et~al.(2020)Wolf, Debut, Sanh, Chaumond, Delangue, Moi, Cistac,
  Rault, Louf, Funtowicz, Davison, Shleifer, von Platen, Ma, Jernite, Plu, Xu,
  Scao, Gugger, Drame, Lhoest, and Rush}]{wolf-etal-2020-transformers}
Wolf, T.; Debut, L.; Sanh, V.; Chaumond, J.; Delangue, C.; Moi, A.; Cistac, P.;
  Rault, T.; Louf, R.; Funtowicz, M.; Davison, J.; Shleifer, S.; von Platen,
  P.; Ma, C.; Jernite, Y.; Plu, J.; Xu, C.; Scao, T.~L.; Gugger, S.; Drame, M.;
  Lhoest, Q.; and Rush, A.~M. 2020.
\newblock Transformers: State-of-the-Art Natural Language Processing.
\newblock In \emph{Proceedings of the 2020 Conference on Empirical Methods in
  Natural Language Processing: System Demonstrations}, 38--45. Online:
  Association for Computational Linguistics.

\bibitem[{Yadan(2019)}]{Yadan2019Hydra}
Yadan, O. 2019.
\newblock Hydra - A framework for elegantly configuring complex applications.
\newblock Github.

\bibitem[{Yates et~al.(2007)Yates, Banko, Broadhead, Cafarella, Etzioni, and
  Soderland}]{yates-etal-2007-textrunner}
Yates, A.; Banko, M.; Broadhead, M.; Cafarella, M.; Etzioni, O.; and Soderland,
  S. 2007.
\newblock {T}ext{R}unner: Open Information Extraction on the Web.
\newblock In \emph{Proceedings of Human Language Technologies: The Annual
  Conference of the North {A}merican Chapter of the Association for
  Computational Linguistics ({NAACL}-{HLT})}, 25--26. Rochester, New York, USA:
  Association for Computational Linguistics.

\bibitem[{Zeng et~al.(2014)Zeng, Liu, Lai, Zhou, and Zhao}]{zeng2014relation}
Zeng, D.; Liu, K.; Lai, S.; Zhou, G.; and Zhao, J. 2014.
\newblock Relation classification via convolutional deep neural network.
\newblock In \emph{Proceedings of COLING 2014, the 25th international
  conference on computational linguistics: technical papers}, 2335--2344.

\bibitem[{Zhan and Zhao(2020)}]{zhan2020span}
Zhan, J.; and Zhao, H. 2020.
\newblock Span model for open information extraction on accurate corpus.
\newblock In \emph{Proceedings of the AAAI Conference on Artificial
  Intelligence}, volume~34, 9523--9530.

\end{thebibliography}
\end{document}